\documentclass[10pt,twocolumn,letterpaper]{article}

\usepackage{cvpr}              

\usepackage[dvipsnames]{xcolor}

\usepackage{amsmath,amsfonts,bm}

\def\eqref#1{equation~\ref{#1}}

\def\1{\bm{1}}

\DeclareMathAlphabet{\mathsfit}{\encodingdefault}{\sfdefault}{m}{sl}
\SetMathAlphabet{\mathsfit}{bold}{\encodingdefault}{\sfdefault}{bx}{n}

\DeclareMathOperator*{\argmin}{arg\,min}

\usepackage{algorithm}
\usepackage{algorithmic}
\usepackage{wrapfig}
\usepackage{enumitem}
\usepackage{graphicx}
\usepackage{comment}
\usepackage{cases}
\usepackage{amsmath,amssymb} 
\usepackage{bbm}
\usepackage{csquotes}

\usepackage{multirow}
\usepackage{booktabs}

\usepackage{pifont}

\usepackage{color, colortbl}
\definecolor{Gray}{gray}{0.9}
\definecolor{TextGreen}{rgb}{0.0,0.69,0.31}
\definecolor{babyblue}{rgb}{0.54, 0.81, 0.94}
\definecolor{firebrick}{rgb}{0.7, 0.13, 0.13}
\definecolor{flame}{rgb}{0.89, 0.35, 0.13}

\newcommand*{\twoelementtable}[3][l]%
{%
    \begin{tabular}[t]{@{}#1@{}}%
        #2\tabularnewline
        #3%
    \end{tabular}%
}

\usepackage{array}
\usepackage{tabularx}

\usepackage{mathtools}

\usepackage{textcomp}
\definecolor{deemph}{gray}{0.6}

\definecolor{baselinecolor}{gray}{.9}

\newlength\savewidth

\definecolor{cvprblue}{rgb}{0.21,0.49,0.74}
\usepackage[pagebackref,breaklinks,colorlinks,citecolor=cvprblue]{hyperref}

\def\paperID{xxxx} 
\def\confName{CVPR}
\def\confYear{2024}

\title{Learning to Prompt Segment Anything Models}

\author{Jiaxing Huang\textsuperscript{\rm 1}\thanks{indicates equal contribution.} , Kai Jiang\textsuperscript{\rm 1}\footnotemark[1] ,  Jingyi Zhang\textsuperscript{\rm 1}, Han Qiu\textsuperscript{\rm 1} \\ Lewei Lu\textsuperscript{\rm 2}, Shijian Lu\textsuperscript{\rm 1}\thanks{corresponding author.} , Eric Xing\textsuperscript{\rm 3}\textsuperscript{\rm 4} \\
\textsuperscript{\rm 1} S-lab, School of Computer Science and Engineering, Nanyang Technological University\\
\textsuperscript{\rm 2} SenseTime Research\\
\textsuperscript{\rm 3} School of Computer Science, Carnegie Mellon University, USA\\
\textsuperscript{\rm 4} Mohamed bin Zayed
University of Artificial Intelligence, Abu Dhabi, UAE\\
{\tt\small \{Jiaxing.Huang, Shijian.Lu\}@ntu.edu.sg}
}

\begin{document}

\maketitle

\begin{abstract}
Segment Anything Models (SAMs) like SEEM and SAM have demonstrated great potential in learning to segment anything.
The core design of SAMs lies with “Promptable Segmentation”, which takes a handcrafted prompt as input and returns the expected segmentation mask.
SAMs work with two types of prompts including spatial prompts (e.g., points) and semantic prompts (e.g., texts), which work together to prompt SAMs to segment anything on downstream datasets.
Despite the important role of prompts, how to acquire suitable prompts for SAMs is largely under-explored.
In this work, we examine the architecture of SAMs and identify two challenges for learning effective prompts for SAMs. To this end, we propose spatial-semantic prompt learning (SSPrompt) that learns effective semantic and spatial prompts for better SAMs.
Specifically, SSPrompt introduces spatial prompt learning and semantic prompt learning, which optimize spatial prompts and semantic prompts directly over the embedding space and selectively leverage the knowledge encoded in pre-trained prompt encoders.
Extensive experiments show that SSPrompt achieves superior image segmentation performance consistently across multiple widely adopted datasets.  Codes will be released.
\end{abstract}

\section{Introduction}

Recently, Segment Anything Models (SAMs), such as Segment Everything Everywhere Model (SEEM)~\citep{zou2023segment} and Segment Anything Model (SAM)~\citep{kirillov2023segment}, have achieved striking image segmentation performance over various downstream datasets~\citep{cordts2016cityscapes,zhou2017scene}, demonstrating their great potential in learning to segment anything. 
The core design lies with ``Promptable Segmentation'', i.e., SAMs take handcrafted prompts as inputs and return expected segmentation masks.
Generally, SAMs work with two types of prompts including spatial prompts (e.g., points or bounding boxes represented by 2D coordinates) and semantic prompts (e.g., free-form texts represented by word tokens), which work together to prompt SAMs to identify and segment anything in images.
However, directly using default prompts (i.e., raw class names as the semantic prompts and a grid of points as the spatial prompts) for every downstream dataset is usually sup-optimal, and how to acquire suitable prompts for SAMs is a non-trivial task as a slight modification of prompts could lead to very different segmentation outcome.

\begin{figure*}[t]
\centering
\includegraphics[width=0.98\linewidth]{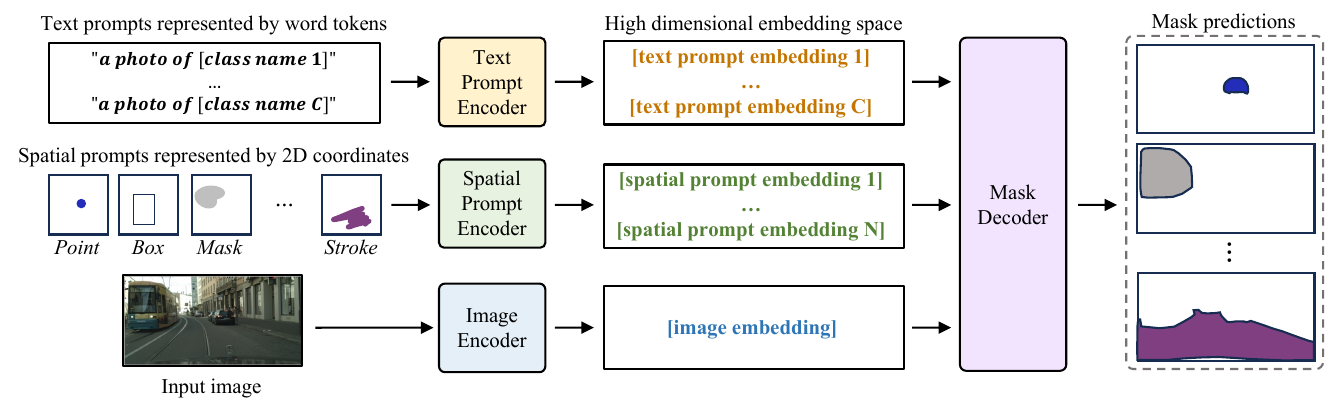}
\caption{
The architecture of Segment Anything Models (SAMs). SAMs~\citep{kirillov2023segment,zou2023segment} consist of three core parts: (1) a large \textit{Image Encoder} that encodes input images into image embeddings; (2) prompt encoders including a large \textit{Text Prompt Encoder} that encodes text tokens into text prompt embeddings and a lightweight \textit{Spatial Prompt Encoder} that encodes 2D spatial coordinates into spatial prompt embeddings; and (3) a lightweight \textit{Spatial Prompt Encoder} that predicts the expected segmentation masks based on the image and prompt embeddings.
}
\label{fig:introduction}
\end{figure*}

By examining the architecture of SAMs in Figure~\ref{fig:introduction}, we identify two challenges of learning effective prompts for SAMs:
(1) \textit{Limited Search Space in Spatial Prompt Learning.} SAMs take XY coordinates in images as spatial prompts, but optimizing such spatial prompts in low-dimensional space (i.e., two dimensions in XY coordinate system) suffers from the limited search space~\citep{koppen2000curse,zimek2012survey} which could lead to sub-optimal spatial prompts.
(2) \textit{Side Effects from Text Prompt Encoder.}
Text prompt encoders in SAMs (e.g., CLIP in SAM and UniCL/Florence in SEEM) are largely pre-trained with object-centric image-text data, where the text data is dominated by the description of foreground objects, leading to well-learnt foreground text knowledge but relatively poorly-learnt background text knowledge.
Consequently, learning semantic prompts with such text prompt encoders can benefit from the well-learnt text knowledge, but may also suffer from the side effects from the poorly-learnt text knowledge.

In this work, we strive for an effective prompt learning technique for SAMs by addressing the above two issues, aiming to acquire optimal spatial and semantic prompts for downstream segmentation datasets with few-shot data.
Considering the architecture of SAMs shown in Figure~\ref{fig:introduction}, we argue that one effective manner to learn prompts for SAMs is by optimizing prompts directly on the embedding space\footnote{Following SAM~\citep{kirillov2023segment} and SEEM~\citep{zou2023segment}, in this paper, “embedding” refers to the representation after the encoder. And ``text tokens'' refer to the text representation before the text encoder.}. 
Intuitively, optimizing spatial prompts directly on the embedding space could relax limited search space, because embedding space is high-dimensional (e.g., 512D) and has much larger search space as compared with 2-dimensional XY coordinate space.
Regarding the Side Effects from Text Prompt Encoder, we argue that the knowledge in text encoder should be utilized selectively so as to benefit from its well-learnt knowledge and concurrently mitigate potential negative effects from its poorly-learnt knowledge.

\begin{figure*}[t]
\centering
\includegraphics[width=0.98\linewidth]{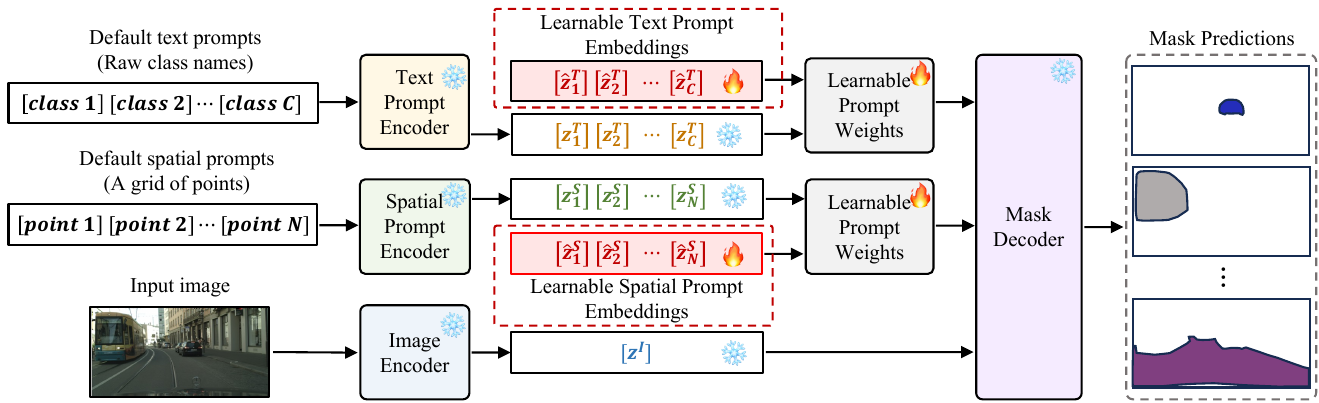}
\caption{
\textbf{The framework of semantic-spatial prompt learning (SSPrompt).}
SSPrompt optimizes spatial and semantic prompts directly on the embedding space and selectively leverages the knowledge encoded in prompt encoders: it employs learnable weights to weight the default prompt embeddings ($\{{z}^{S}_{n}\}_{n=1}^{N}$ and $\{{z}^{T}_{c}\}_{c=1}^{C}$) and fuses the weighted embeddings with the learnable prompt embeddings (i.e., $\{\hat{z}^{S}_{n}\}_{n=1}^{N}$ and $\{\hat{z}^{T}_{c}\}_{c=1}^{C}$) to acquire new prompts.
During training, only the \textit{Learnable Prompt Embeddings} and the \textit{Learnable Prompt Embeddings} are updated (marked by \textcolor{flame}{Flame}), while all rest are frozen (marked by \textcolor{babyblue}{Snowflake}).
}
\label{fig:method}
\end{figure*}

To this end, we design spatial-semantic prompt learning (SSPrompt) that introduces spatial prompt learning (SpaPrompt) and semantic prompt learning (SemPrompt) for learning effective prompts for SAMs, as illustrated in Figure~\ref{fig:method}.
For semantic prompt learning, SemPrompt employs learnable weights to weight the default semantic prompt embeddings (encoded by fixed \textit{Text Prompt Encoder}) and then fuses the weighted embeddings with a set of \textit{Learnable Semantic Prompt Embeddings} to acquire new semantic prompts.
Intuitively, SemPrompt 1) is efficient as its optimization only involves the embeddings encoded by the large text prompt encoder instead of the text prompt encoder itself, and 2) can mitigate potential side effects from the text prompt encoder by introducing learnable weights to selectively leverage the knowledge encoded in the encoder (i.e., the default semantic prompt embeddings encoded by the encoder).
For spatial prompt learning, SpaPrompt employs learnable weights to weight the default spatial prompt embeddings (encoded by the fixed spatial prompt encoder) and fuses the weighted embeddings with a set of learnable spatial prompt embeddings to acquire new spatial prompts.
In this way, SpaPrompt relaxes the limited search space by optimizing spatial prompts on high-dimensional embedding space. Similar to SemPrompt, SpaPrompt can selectively utilize the knowledge encoded in spatial prompt encoder.

The contributions of this work can be summarized in three major aspects. First, we identify two challenges in prompt learning in SAMs and investigate how to tackle them for the first time to the best of our knowledge.
Second, we design spatial-semantic prompt learning which directly optimizes spatial and semantic prompts in the embedding space and selectively exploit the knowledge encoded in prompt encoders, ultimately learning effective prompts for SAM using few-shot data only.
Third, extensive experiments show that the proposed method achieves state-of-the-art performances consistently over multiple widely adopted segmentation datasets.

\section{Related Work}

\textbf{Segment Anything Models} (SAMs) have recently demonstrated great potential in learning to segment
anything~\citep{kirillov2023segment,zou2023segment}, which achieve striking image segmentation~\cite{long2015fully,zhao2017pyramid,chen2017deeplab} performance over various downstream datasets.
To our knowledge, the recent breakthroughs of SAMs, particularly SAM~\citep{kirillov2023segment} and SEEM~\citep{zou2023segment}, are largely driven by the advanced design called ``Promptable Segmentation'', i.e., SAMs take a handcrafted prompt as input and return the expected segmentation mask.
Generally, SAMs involve two types of prompts including semantic prompts (e.g., free-form texts) and spatial prompts (e.g., points or bounding boxes), which provide semantic and spatial information respectively and together prompt segmentation models to identify and segment anything in images~\citep{kirillov2023segment,zou2023segment}.
On the other hand, how to acquire suitable prompts for SAMs is a non-trivial task but largely under-explored.
In this work, we focus on investigating how to learn effective prompts for SAMs using few-shot data, aiming to facilitate the deployment of SAMs for task-specific or domain-specific downstream datasets.

\textbf{Prompt Learning} aims to adapt foundation models towards downstream tasks by optimizing prompts using few-shot data. In recent years, prompt learning has been extensively studied for NLP foundation models~\citep{zhong2021factual,li2021prefix,lester2021power,liu2023pre} and image classification foundation models (CFMs)~\citep{zhou2022learning,parisot2023learning,zang2022unified,shen2022multitask,khattak2022maple,xing2022class,gao2021clip,zhang2021tip,pantazis2022svl,udandarao2022sus,kahana2022improving,peng2022sgva,zhang2021vt,guo2022calip,yu2023task,pratt2022does,menon2022visual,wortsman2022robust}.
Specifically, for NLP foundation models that generally work in a question answering manner, various prompt learning methods have been introduced to learn effective context text tokens to append and improve the raw questions, such as text mining/paraphrasing~\citep{jiang2020can}, gradient-based searching~\citep{shin2020autoprompt}, continuous text token optimization~\citep{zhong2021factual,li2021prefix,lester2021power,liu2023pre}.
On the other hand, for CFMs that classify images based on class names, a variety of prompt learning methods~\citep{zhou2022learning,zhou2022conditional,parisot2023learning,ma2022understanding,bulat2022language,lu2022prompt,derakhshani2022variational,zhu2022prompt,he2022cpl,chen2022prompt,sundualcoop,guo2022texts,ding2022prompt,rao2022denseclip,huang2022unsupervised,shutest,yao2023visual,wu2023protect,bahng2022exploring,rong2023retrieval} have been proposed to learn effective context text tokens to append and improve the raw class names, such as continuous text token optimization~\citep{zhou2022learning}, conditional text token optimization~\citep{parisot2023learning}, etc.
Different from previous works that focus on image classification foundation models and optimize text tokens to prompt text encoder, we examine the architecture of Segment Anything Models (SAMs) and propose a more efficient and effective prompt learning method for SAMs. The method directly
optimizes spatial and semantic prompts in the embedding space and selectively exploits the knowledge encoded in prompt encoder \footnote{Following SAM~\citep{kirillov2023segment} and SEEM~\citep{zou2023segment}, in this paper, “embedding” refers to the representation after the encoder. And ``text tokens'' refer to the text representation before the text encoder.}, ultimately learning effective prompt embeddings to prompt the mask decoder.

\section{Method}
In this section, we first introduce the background of Segment Anything Models (Section~\ref{sec:31}) and revisit the prompt learning methods of image classification foundation models (Section~\ref{sec:32}). Then, we elaborate our proposed prompt learning method for Segment Anything Models (Section~\ref{sec:33}).

\subsection{Preliminaries of Segment Anything Models}
\label{sec:31}

Segment Anything Models (SAMs)~\citep{zou2023segment,kirillov2023segment} learn to segment anything by introducing a new ``Promptable Segmentation'' scheme, where the segmentation model predicts the expected segmentation mask for a given prompt.
In this way, SAMs could segment anything (e.g., any objects and background stuff) given proper prompts.
Besides, SAMs enable ``interactive segmentation''~\cite{mcguinness2010comparative,sofiiuk2020f,mortensen1998interactive} that helps scale up the segmentation training data by using a data engine with model-in-the-loop annotating, which in turn facilitates training more powerful SAMs for better ``interactive segmentation''~\cite{zhang2023personalize,liu2023matcher}.

Specifically, SAMs~\citep{zou2023segment,kirillov2023segment} consist of three core parts: (1) an image encoder that encodes images into image embeddings; (2) a text prompt encoder and a spatial prompt encoder, which respectively encode text prompts and spatial prompts into prompt embeddings; (3) a mask decoder that returns the expected segmentation mask based on the image and prompt embeddings.

Given an input image $x^{I} \in \mathbb{R}^{H \times W \times 3}$ and a set of prompts (e.g., a spatial prompt $x^{S}$ and a text prompt $x^{T}$), SAMs first employ the image encoder $Encoder^{I}$, the spatial prompt encoder $Encoder^{S}$ and the text prompt encoder $Encoder^{T}$ to encode them into $D$-dimensional embeddings: $z^{I} = Encoder^{I}(x^{I})$, $z^{S} = Encoder^{S}(x^{S})$ and $z^{T} = Encoder^{T}(x^{T})$, respectively.
Then, given the encoded image and prompts, the mask decoder of SAMs predicts the expected segmentation mask:
\begin{equation}
    (m, c) = Decoder(z^{I}|z^{S}, z^{T}),
    \label{eq_sfm}
\end{equation}
where $m$ stands for a predicted binary segmentation mask and $c$ denotes the predicted confidence score of $m$.
In SEEM~\citep{zou2023segment}, $c$ stands for the probability of mask $m$ belonging to the category denoted by text $x^{T}$.
In SAM~\citep{kirillov2023segment}, when prompted by text prompts, $c$ can also denote the probability of mask $m$ belonging to the category denoted by text $x^{T}$. When prompted by spatial prompts, $c$ in SAM is class-agnostic and only denotes the quality of predicted mask $m$.

Note, for spatial prompt $x^{S}$, we mainly consider the format of point, i.e., $x^{S} = (h, w)$ ($h \in (0, H)$ and $w \in (0, W)$ where $H$ and $W$ denote image height and image width respectively), because all other formats of spatial prompts can be represented in terms of points, e.g., the bounding box can be denoted by two corner points and the  coarse mask can be denoted by a set of points.

\textbf{Zero-shot Cross-Dataset Inference.} Given an image $x^{I}$ and a set of default prompts (i.e., raw
class names $X^{T}_{\text{default}} = \{x^{T}_{c}\}_{c=1}^{C}$ as the semantic prompts and a grid of points $X^{S}_{\text{default}} = \{x^{S}_{n}\}_{n=1}^{N}$ as the spatial prompts), SAMs~\citep{zou2023segment} can predict a set of segmentation masks for $x^{I}$:
\begin{equation}
    (M, C) = Decoder(z^{I}|Z^{S}_{\text{default}}, Z^{T}_{\text{default}}),
    \label{eq_sfm_inference}
\end{equation}
where $Z^{S}_{\text{default}} = Encoder^{S}(X^{S}_{\text{default}})$ and $Z^{T}_{\text{default}} = Encoder^{T}(X^{T}_{\text{default}})$.

On the other hand, directly using default prompts for every downstream dataset is
usually sup-optimal, and how to acquire suitable prompts for SAMs is a non-trivial task but largely under-explored. In this work, we focus on investigating how to learn effective prompts for SAMs using few-shot data.

\subsection{A Revisit of Prompt Learning}
\label{sec:32}

Prompt Learning aims to adapt a foundation model towards downstream tasks by optimizing the prompts using few-shot data. In recent years, various prompt learning methods have been proposed for image classification foundation models (CFMs)~\citep{radford2021learning,parisot2023learning}.
The core idea of CFM prompt leaning methods is to learn effective context text tokens to append and improve the raw class names, for better prompting the text encoder.
Specifically, CFM prompt leaning methods, such as CoOp, introduce $M$ learnable context text tokens, i.e., $x^{T}_{\text{context}} = \{x^{T}_{1}, x^{T}_{2}, ..., x^{T}_{M}\}$, to model the context of each raw class name $x^{T} \in X^{T}_{\text{default}}$, such that the text prompts become $X^{T}_{CoOp} = \{X^{T}_{\text{default}}, X^{T}_{\text{context}}\}$.
Given an image $x^{I}$, the text prompt $X^{T}_{CoOp}$ and a CFM consisting of an image encoder $Encoder^{I}$ and a text encoder $Encoder^{T}$, the image classification prediction can be formulated by:
\begin{equation}
    c = Encoder^{I}(x^{I}) \cdot Encoder^{T}(\{X^{T}_{\text{default}}, X^{T}_{\text{context}}\}),
    \label{eq_cfm_inference}
\end{equation}
where `$\cdot$' denotes the inner (dot) product that measures the similarity between the image embedding and text embeddings. $X^{T}_{\text{default}}$ and $X^{T}_{\text{context}}$ are concatenated categorically before being fed into the text encoder.

To adapt CFMs to a downstream dataset, an image classification loss can be employed as the learning objective to optimize $X^{T}_{\text{context}}$ over few-shot data while keeping all other modules unchanged.

Different from previous works that focus on image classification foundation models and optimize text tokens to prompt text encoder, we examine the architecture of Segment Anything Models (SAMs) and propose a more efficient and effective prompt learning method for SAMs. Specifically, we optimize spatial and semantic prompts in the embedding space and selectively exploit the knowledge encoded in prompt encoder, ultimately learning effective prompt embeddings to prompt the mask decoder.

\subsection{Spatial-Semantic Prompt Learning}
\label{sec:33}

We focus on prompt learning for SAMs using few-shot data.
By examining the architecture of SAMs, we identify two challenges of learning effective prompts for SAMs: (1) \textit{Limited Search Space in Spatial Prompt Learning.} (2) \textit{Side Effects from Text Prompt Encoder.}
We propose to tackle the two challenges by 1) directly optimizing prompts on the embedding space and 2) selectively leveraging the knowledge encoded in the pretrained prompt encoder. To this end, we design spatial-semantic prompt learning (SSPrompt) that introduces spatial prompt
learning (SpaPrompt) and semantic prompt learning (SemPrompt), as illustrated in Figure~\ref{fig:method}.
The two prompt learning methods complement each other by capturing spatial and semantic information respectively, which together learn effective spatial and semantic prompts for SAMs.

\textbf{Spatial prompt learning} (SpaPrompt) optimizes spatial prompts directly on the embedding space and selectively leverages the knowledge encoded in the pretrained spatial prompt encoder: it employs learnable
weights to weight the default spatial prompt embeddings (encoded by the fixed spatial prompt encoder) and fuses the weighted embeddings with a set of learnable spatial prompt embeddings to acquire
new spatial prompts.
In this way, SpaPrompt learns effective spatial prompts for SAMs with two desirable features: 1) It relaxes the limited search space by optimizing spatial prompts directly on high-dimensional embedding space that has larger search space (e.g., 512 dimensions) than 2D coordinate space (i.e., 2 dimensions); 2) Similar to SemPrompt (mentioned in latter paragraphs), SpaPrompt can selectively utilize the knowledge encoded in the pretrained spatial prompt encoder.

Let $\hat{Z}^{S} = \{\hat{z}^{S}_{n}\}_{n=1}^{N}$ denote $N$ learnable spatial embeddings, where $\hat{z}^{S}_{n} \in \mathbb{R}^{D}$ and $D$ denotes the demension of embedding, and $\hat{W}^{S} = \{\hat{w}^{S}_{n}\}_{n=1}^{N}$ denote $N$ learnable weights, where $\hat{w}^{S}_{n} \in [0, 1]$.
The new spatial prompt $Z^{S}_{\text{SpaPrompt}}$ can be obtained by applying $\hat{W}^{S}$ to weight the default spatial prompt embeddings $Z^{S}_{\text{default}}$ and fusing the weighted embeddings with $\hat{Z}^{S}$: 
\begin{equation}
    Z^{S}_{\text{SpaPrompt}} = \{\hat{w}^{S}_{n} \hat{z}^{S}_{n} + (1-\hat{w}^{S}_{n})z^{S}_{n}\}_{n=1}^{N},
    \label{eq_spa}
\end{equation}
where $\hat{z}^{S}_{n} \in \hat{Z}^{S}, \ z^{S}_{n} \in Z^{S}_{\text{default}} \ \text{and} \ \hat{w}^{S}_{n} \in \hat{W}^{S}$.

Given an image $x^{I}$ and the new spatial prompt $Z^{S}_{\text{SpaPrompt}}$, SAMs predict a set of segmentation masks:
\begin{equation}
    (M, C) = Decoder(z^{I}| Z^{S}_{\text{SpaPrompt}}, Z^{T}_{\text{default}}),
    \label{eq_spa_inference}
\end{equation}
where we can employ a segmentation loss to optimize $Z^{S}_{\text{SpaPrompt}}$ to find the best spatial prompts for SAMs with respect to each downstream dataset.
Note, during training, we only update the learnable embeddings $\hat{Z}^{S}$ and the learnable weights $\hat{W}^{S}$ to optimize $Z^{S}_{\text{SpaPrompt}}$, while all other modules have been frozen as illustrated in Figure~\ref{fig:method}.

\textbf{Semantic prompt learning} (SemPrompt) 

optimizes semantic prompts directly on the embedding space and selectively leverages the knowledge encoded in the pretrained text prompt encoder: it employs learnable weights to weight the default semantic prompt embeddings (encoded by the fixed text prompt encoder) and fuses the weighted embeddings with a set of learnable semantic prompt embeddings to acquire
new semantic prompts.
SemPrompt learns semantic prompts for SAMs with two desirable features: 1) It 
is efficient as its optimization only involves the embeddings encoded by the large text prompt encoder instead of the text prompt encoder itself;
2) Its design of learnable weights allows to selectively leverage the semantic knowledge in the default semantic prompt embeddings and the learnable semantic prompt embeddings, which helps capture complementary knowledge, i.e., the former is encoded by the fixed text prompt encoder (pre-trained on large-scale data) and captures general semantic knowledge, while the latter is optimized and learnt from the downstream data and largely captures task-specific and domain-specific semantic knowledge.

Let $\hat{Z}^{T} = \{\hat{z}^{T}_{c}\}_{c=1}^{C}$ denote $C$ learnable semantic embeddings, where $\hat{z}^{T}_{c} \in \mathbb{R}^{D}$ and $D$ denotes the dimension of embedding, and $\hat{W}^{T} = \{\hat{w}^{T}_{c}\}_{c=1}^{C}$ denote $C$ learnable weights, where $\hat{w}^{T}_{c} \in [0, 1]$.
The new semantic prompt $Z^{T}_{\text{SemPrompt}}$ can be obtained by applying $\hat{W}^{T}$ to weight the default semantic prompt embeddings $Z^{T}_{\text{default}}$ and fusing the weighted embeddings with $\hat{Z}^{T}$: 
\begin{equation}
    Z^{T}_{\text{SemPrompt}} = \{\hat{w}^{T}_{c} \hat{z}^{T}_{c} + (1-\hat{w}^{T}_{c})z^{T}_{c}\}_{c=1}^{C}, 
    \label{eq_sem}
\end{equation}
where $\hat{z}^{T}_{c} \in \hat{Z}^{T}, \ z^{T}_{c} \in Z^{T}_{\text{default}} \ \text{and} \ \hat{w}^{T}_{c} \in \hat{W}^{T}$.

Given an image $x^{I}$ and new semantic prompt $Z^{T}_{\text{SemPrompt}}$, SAMs predict a set of segmentation masks:
\begin{equation}
    (M, C) = Decoder(z^{I}| Z^{S}_{\text{default}}, Z^{T}_{\text{SemPrompt}}),
    \label{eq_sem_inference}
\end{equation}
where we can employ a segmentation loss to optimize $Z^{T}_{\text{SemPrompt}}$ to find the best semantic prompts for SAMs with respect to each downstream dataset. 
Note, during training, we only update the learnable embeddings $\hat{Z}^{T}$ and the learnable weights $\hat{W}^{T}$ to optimize $Z^{T}_{\text{SemPrompt}}$, while all other modules have been frozen as illustrated in Figure~\ref{fig:method}.

\textbf{Spatial-semantic prompt learning} (SSPrompt) combines spatial prompt learning and semantic prompt learning, aiming for leveraging the synergy of spatial and semantic information to better prompt Segment Anything Models.
Given an image $x^{I} \in X^{I}$ and its segmentation annotation $y^{I} \in Y^{I}$, the new spatial prompt $Z^{S}_{\text{SpaPrompt}}$ from Eq.~\ref{eq_spa} and the new semantic prompt $Z^{T}_{\text{SemPrompt}}$ from Eq.~\ref{eq_sem}, SSPrompt can be formulated as:
\begin{align}
    \{M, C\} = Decoder(z^{I}| Z^{S}_{\text{SpaPrompt}}, Z^{T}_{\text{SemPrompt}}), \\
    \argmin_{\{\hat{Z}^{S}, \hat{W}^{S}, \hat{Z}^{T}, \hat{W}^{T}\}} \frac{1}{|X^{I}|}  \sum_{x^{I} \in X^{I}}{\mathcal{L}_{seg} (\{M, C\}, y^{I})},
    \label{eq_ssprompt_inference}
\end{align}
where $\mathcal{L}_{seg}$ denotes a standard segmentation loss function and $z^{I} = Encoder^{I}(x^{I})$.
Note we initialize $Z^{S}_{\text{default}}$ and $Z^{T}_{\text{default}}$ as in Eq.~\ref{eq_sfm_inference} before training such that the training process of SSPrompt will not involve the spatial prompt encoder and the large text prompt encoder.

\section{Experiments}

\begin{table*}[h]
\centering
\resizebox{\textwidth}{!}{
    \begin{tabular}	{l r r l}
    \toprule
    Dataset  & Classes & Images & Description \\
    \midrule
    Cityscapes~\citep{cordts2016cityscapes} &19 &5,000  & Street scene images ($\sim$1080p) from European cities under good weather conditions.\\
    BDD100K~\citep{yu2020bdd100k} &19 &10,000 & Street scene images ($\sim$720p) from American cities under various weather conditions.\\
    Mapillary~\citep{neuhold2017mapillary} &19 &25,000  & Street scene images from all over the world with high resolutions, e.g., $4000 \times 5000$\\
    ADE20K~\citep{zhou2017scene} & 150 &27,574 & A large-scale dataset with 20K+ scene-centric images and 150 semantic categories.\\ 
    Pascal Context~\citep{mottaghi2014role} &59 &10,103  & An extension of the PASCAL VOC 2010 detection challenge with pixel-wise labels. \\
    ACDC~\citep{sakaridis2021acdc} & 19 & 4,006   & An adverse conditions dataset with fog, nighttime, rain, and snow conditions. \\
	\bottomrule
	\end{tabular}
}
\caption
{Datasets used to benchmark prompt learning for segment anything models.
}
\label{table:datasets}
\end{table*}

\subsection{Datasets}

We benchmark our SSPrompt extensively over 6 widely used image segmentation datasets with pixel-wise annotations. As Table~\ref{table:datasets} shows, the 6 datasets have rich diversity, spanning from street scene data that include high-resolution images captured in different cities and under various daytimes, weathers and seasons, to category-rich data that cover 59 and 150 semantic categories.
We did not include COCO dataset in experiments as it has been used in SAMs pre-training~\cite{zou2023segment}.

\subsection{Implementation Details}

We conduct experiments with two vision backbones including Focal-Tiny~\citep{yang2022focal} and Davit-Large~\citep{ding2022davit}.
In training, we employ SGD optimizer~\cite{loshchilov2017decoupled} with a weight decay of $1e-4$, and set the base learning rate as $1e-3$ which is adjusted with a polynomial learning rate schedule with a power of $0.9$.
We use $1$ GPU with batch size $2$ for Cityscapes, BBD and ACDC, and $4$ GPUs with batch size $8$ for large datasets Mapillary, ADC20K and PASCAL Context. Our prompt learning method introduces very limited computation overhead, as illustrated in Table~\ref{table:efficiency} and appendix.
We set the shorter side of input images at $512$ and employ random flip as data augmentation.
The number of semantic prompts $C$ is set as the number of categories of each downstream dataset. Following~\citet{zou2023segment}, we set the number of spatial prompts $N$ as 100 and the dimension of embedding $D$ as $512$.  
Following~\citet{chen2017deeplab}, we employ cross-entropy loss~\cite{de2005tutorial} for semantic segmentation. For instance segmentation and panoptic segmentation, we use multi-category cross-entropy loss for class prediction training and binary cross-entropy loss for mask prediction training. 

\begin{table*}[h]
\begin{center}
\begin{tabular}{lcccccc}
\toprule
\multicolumn{6}{c}{\textbf{Experiments with Tiny Vision Backbone}} \\\midrule
Method &  Cityscapes
& BDD100K
& Mapillary
& ADE20K
& PASCAL Context
\\
\midrule
SEEM-T~\citep{zou2023segment} &39.2 &37.4 &42.1 &14.6 &45.1\\
CoOp~\citep{zhou2022learning} & 50.1 &41.6 &43.3 &17.6 &45.9\\
LOCN~\citep{parisot2023learning} & 51.5 &42.6 &44.2 &19.3 &46.6\\
\rowcolor{gray!16} SSPrompt (Ours) &55.2 &47.1 &49.5 &23.2 &51.2
\\
\bottomrule
\toprule
\multicolumn{6}{c}{\textbf{Experiments with Large Vision Backbone}} \\\midrule
Method &  Cityscapes
& BDD100K
& Mapillary
& ADE20K
& PASCAL Context  
\\
\midrule
SEEM-L~\citep{zou2023segment} &49.3	& 44.6	& 47.9	& 15.2	& 37.1\\
CoOp~\citep{zhou2022learning} &51.2  &45.2 &52.0 &18.1 &47.4\\
LOCN~\citep{parisot2023learning} & 52.7 &45.7 &53.2 &19.7 &48.9\\
\rowcolor{gray!16} SSPrompt (Ours) &57.1 &49.5 &56.2 &25.6 &55.3
\\
\bottomrule
\end{tabular}
\end{center}
\caption{Prompt learning of Segment Anything Models on common datasets. The experiments are conducted on semantic segmentation (in mIoU), where 16-shot data are used (i.e., 16 labelled images for each class) for each dataset.
}
\label{tab:all_seg}
\end{table*}

\subsection{Prompt Learning for SAMs on Common Datasets}
Table~\ref{tab:all_seg} reports the image segmentation results on 5 widely-used common datasets.
It can be seen that our SSPrompt achieves superior prompt learning performance consistently over various segmentation datasets.
The superior performance is largely attributed to our two prompt learning designs that effectively address the two identified challenges in prompt learning for SAMs.
Besides, it is expected that the large model SEEM-L should outperform the small model SEEM-T while SEEM-L performs unexpectedly not well on PASCAL Context dataset, where all prompt learning methods improve the performance while our SSPrompt brings the most substantial performance gain, showing that SSPrompt can well handle the occasional failures of SAMs.

\begin{table*}[h]
\begin{center}
\begin{tabular}{lccccc}
\toprule
\multicolumn{6}{c}{\textbf{Experiments with Tiny Vision Backbone}} \\\midrule
Method &\multicolumn{1}{c}{Foggy Condition}  &\multicolumn{1}{c}{Night Condition} 
&\multicolumn{1}{c}{Rain Condition}
&\multicolumn{1}{c}{Snow Condition} &\multicolumn{1}{c}{Mean}
\\
\midrule
SEEM-T~\citep{zou2023segment} &34.6	&26.2	&33.1	&35.8	&32.4 \\
CoOp~\citep{zhou2022learning} &36.7 &28.6 &33.5 &36.4 &33.8 \\
LOCN~\citep{parisot2023learning} &40.1 &29.1 &34.1 &36.6 &35.0 \\
\rowcolor{gray!16} SSPrompt (Ours) &47.5 &32.1 &39.9 &43.1 &40.6\\
\bottomrule
\toprule
\multicolumn{6}{c}{\textbf{Experiments with Large Vision Backbone}} \\\midrule
Method &\multicolumn{1}{c}{Foggy Condition}  &\multicolumn{1}{c}{Night Condition} 
&\multicolumn{1}{c}{Rain Condition}
&\multicolumn{1}{c}{Snow Condition} &\multicolumn{1}{c}{Mean}
\\
\midrule
SEEM-L~\citep{zou2023segment} &48.1	&32.0	&47.4	&45.0 &43.1\\
CoOp~\citep{zhou2022learning} &52.2 &33.5 &48.2 &45.6 &44.9\\
LOCN~\citep{parisot2023learning} &53.7 &33.8 &49.5 &45.9 &45.7\\
\rowcolor{gray!16} SSPrompt (Ours) &57.7 &37.8 &54.5 &49.5 &49.9
\\
\bottomrule
\end{tabular}
\end{center}
\caption{Prompt learning of Segment Anything Models on adverse-condition dataset, i.e., ACDC~\citep{sakaridis2021acdc}. The experiments are conducted on semantic segmentation (in mIoU), where 16-shot data are used (i.e., 16 labelled images for each class) for each condition.
}
\label{tab:adverse_seg}
\end{table*}

\subsection{Prompt Learning for SAMs on Adverse-Condition Dataset}
Table~\ref{tab:adverse_seg} reports the image segmentation results over the adverse-condition dataset, i.e., ACDC~\citep{sakaridis2021acdc}.
We can observe that SSPrompt outperforms the state-of-the-art by large margins consistently over different adverse conditions, demonstrating
its great potential for more robust SAMs by learning effective domain-specific prompts.

\subsection{Discussion}

\textbf{Generalization across different datasets.} We examine the generalization of SSPrompt with respect to image segmentation datasets.
Specifically, we perform extensive evaluations over 6 widely studied common and adverse-condition datasets as described in Table~\ref{table:datasets}.
Experimental results in Tables~\ref{tab:all_seg}-~\ref{tab:adverse_seg} show that SSPrompt achieves superior performance consistently across different types of image data.

\textbf{Generalization across different vision backbones.} We study the generalization of SSPrompt by evaluating it with two vision backbones, including Focal-Tiny~\citep{yang2022focal} and Davit-Large~\citep{ding2022davit}.
Results in Tables~\ref{tab:all_seg}-~\ref{tab:adverse_seg} show that SSPrompt works effectively and consistently over both small and large vision backbones. Note we did not conduct experiments using SAM~\citep{kirillov2023segment} as its version with text prompt encoder is not open-sourced and the released SAM version can only support class-agnostic segmentation.

\begin{table}[ht]
\centering	
\begin{tabular}	{l | c|c|c} 
\toprule
Cityscapes & Sem. Seg &Ins. Seg &Pan. Seg\\
\midrule
SEEM-T & 39.2 &32.7 &32.4 \\
\rowcolor{gray!16} SSPrompt & 55.2 &37.7 &38.0 \\
\bottomrule
\end{tabular}
\caption{
Results on semantic (mIoU), instance (AP50) and panoptic segmentation (PQ).}
\label{table:tasks}
\end{table}	

\textbf{Generalization across different tasks.} 
We also examine the generalization of SSPrompt over different segmentation tasks including semantic segmentation, instance segmentation and panoptic segmentation.
As Table~\ref{table:tasks} shows, SSPrompt improves the performance across all three segmentation tasks consistently. All experiments are conducted under the same setup with 16-shot data.

\begin{table*}[h]
\begin{center}
\resizebox{0.98\linewidth}{!}{
\begin{tabular}{lccccc}
\toprule
\multirow{2}*{Method} & \multicolumn{2}{c}{Spatial Prompt Learning} & \multicolumn{2}{c}{Semantic Prompt Learning}
& \multirow{2}*{mIoU}\\
\cmidrule(lr){2-3}
\cmidrule(lr){4-5}
&  \multicolumn{1}{c}{Learnable Prompt Embedding}  & \multicolumn{1}{c}{Learnable Prompt Weight} &  \multicolumn{1}{c}{Learnable Prompt Embedding}  & \multicolumn{1}{c}{Learnable Prompt Weight}  \\
\midrule
SEEM-T &&& &&39.2\\
\midrule
&\checkmark&&&&46.2\\
&\checkmark&\checkmark&&&49.3\\
&&&\checkmark&& 51.5 \\
&&&\checkmark&\checkmark& 53.1 \\
\rowcolor{gray!16} SSPrompt &\checkmark&\checkmark&\checkmark&\checkmark&55.2\\\bottomrule
\end{tabular}
}
\end{center}
\caption{
Ablation studies of SSPrompt on Cityscapes dataset using 16-shot data.
}
\vspace{-1mm}
\label{table:ablation}
\end{table*}

\textbf{Ablation study.} We conduct ablation studies with Focal-Tiny on Cityscapes as shown in Table~\ref{table:ablation}. We examine how SSPrompt's two core designs, i.e., 1) directly optimizing prompts on embedding space and 2) selectively leveraging the knowledge in prompt encoders, contribute to the overall performance.
As Table~\ref{table:ablation} show, directly optimizing prompts on embedding space (i.e., optimizing learnable prompt embedding and average it with the default prompt embedding) improves the performance clearly, demonstrating its effectiveness on both spatial prompt learning and semantic prompt learning for better image segmentation with SMFs.
In addition, instead of simply averaging, introducing learnable weights to selectively weight and fuse the default prompt embedding and the learnable prompt embedding brings further performance improvements, indicating
that the learnable weights enable more effective usage of prompt encoders' knowledge and help learn better prompts.
Moreover, combining spatial and semantic prompt learning performs the best clearly, demonstrating that the two types of prompt learning methods complement each other by providing orthogonal spatial and semantic information.

\begin{table}[ht]
\centering	
\begin{tabular}	{c | cccc } 
\toprule
\multirow{2}*{SEEM-T} &\multicolumn{4}{c}{SSPrompt} \\
&\cellcolor{gray!16}{16-shot} & 12-shot &8-shot &4-shot\\
\midrule
39.2 & \cellcolor{gray!16}{55.2} &52.6 &50.6 & 50.1   \\
\bottomrule
\end{tabular}
\caption{
Performance (in mIoU) versus number of data. The default is marked in \colorbox{gray!16}{gray}.}
\label{table:shot}
\end{table}

\textbf{Performance versus the number of training data.}
We investigate how the amount of training data affects the performance by reducing it from 16-shot to 4-shot progressively. As shown in Table~\ref{table:shot}, SSPrompt still brings clear performance improvements against the baseline SEEM-T with less training data, showing the effectiveness of SSPrompt on different amounts of training data.

\begin{table}[ht]
\centering	
\begin{tabular}	{l | ccc } 
\toprule
SEEM-T & CoOp &LOCN &\cellcolor{gray!16}{SSPrompt}\\
\midrule
Training Time &87.5	&89.5	&\cellcolor{gray!16}{56.0 \textcolor{TextGreen}{(-36.0\%)}} \\
Training Memory &8.22	&8.22	&\cellcolor{gray!16}{3.82 \textcolor{TextGreen}{(-53.5\%)}} \\
\bottomrule
\end{tabular}
\caption{
Training efficiency comparison in time (ms per image) and memory (GB).}
\label{table:efficiency}
\end{table}

\textbf{Training efficiency comparison.} We analyze training efficiency by comparing prompt learning methods in training time (millisecond per image) and training memory (GB). Table~\ref{table:efficiency} shows the results on ADE20K, indicating that SSPrompt is more efficient in training time and training memory. The superior efficiency is largely because SSPrompt circumvents the large text prompt encoder and requires less computation and memory. More results on other datasets are provided in the appendix.

\begin{figure*}[ht]
\centering
\includegraphics[width=0.98\linewidth]{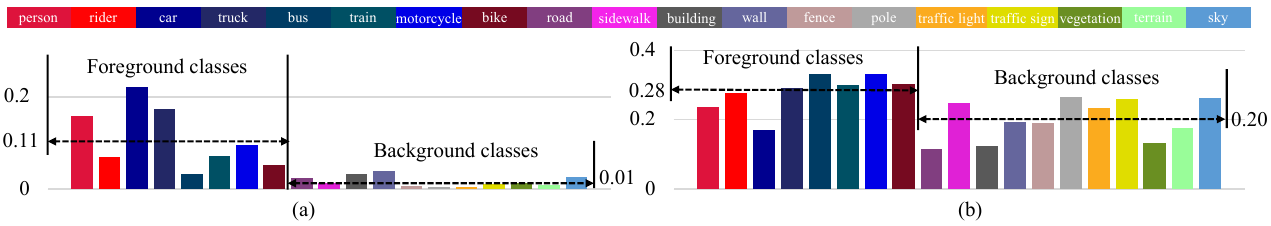}
\\
\caption{(a) Text data statistics (used for text prompt encoder pre-training in SAMs~\citep{zou2023segment,kirillov2023segment}). (b) Learnt weights in semantic prompt learning.
}
\vspace{-2ex}
\label{fig:weight}
\end{figure*}

\textbf{Side Effects from Text Prompt Encoder.} We investigate the 
bias of Text Prompt Encoder and its side effects by comparing the text data statistics used to pre-train it, i.e., the occurrence of each class names in widely-used image-text dataset, LAION~\citep{schuhmann2021laion}. As shown in Figure~\ref{fig:weight} (a), the foreground class names generally occur much more frequently than background class names, which indicates that the text knowledge learnt from these data (i.e., the knowledge encoded in text prompt encoder) would bias toward foreground objects, leading to well-learnt foreground text knowledge but relatively poorly-learnt background text knowledge. Consequently, learning semantic prompts with such text prompt encoders can benefit from the well-learnt text knowledge, but may also suffer from the side effects from the poorly-learnt text knowledge. 
This is aligned with the ablation studies in Table~\ref{table:ablation}, where introducing learnable weights to selectively exploit prompt encoder's knowledge improves the segmentation performance clearly.
In addition, Figure~\ref{fig:weight} (b) visualizes the learnt weights in semantic prompt learning, where background classes are generally assigned with lower weights while foreground classes are often assigned with higher weights, showing that the foreground knowledge in text prompt encoder is more helpful in semantic prompt learning while background knowledge is less helpful.

\begin{table}[ht]
\centering	
\begin{tabular}	{l |c  ccc } 
\toprule
Method &SEEM-T & VSPL &SpaPrompt &\cellcolor{gray!16}{SSPrompt}\\
\midrule
Cityscapes & 39.2 & 41.0 &49.3 & \cellcolor{gray!16}{55.2} \\
\bottomrule
\end{tabular}
\caption{
Comparison with Vanilla Spatial Prompt Learning (VSPL) on 16-shot data in mIoU.}
\label{table:vanilla}
\end{table}

\textbf{Limited Search Space.} We investigate how much the Limited Search Space issue affects learning spatial prompts by implementing Vanilla Spatial Prompt Learning (VSPL) that optimizes spatial prompts in 2D coordinate system. Results in Table~\ref{table:vanilla} show that VSPL does not help much, largely due to the limited search space in VSPL. On the other hand, our SpaPrompt (and SSPrompt) optimizes prompts directly on high-dimensional embedding space, leading to larger search space and clearly improved performance.

Due to the space limit, we provide more dataset details, experiments and discussions in the appendix.

\section{Conclusion}

In this work, we identify two challenges of learning effective prompts for SAMs by examining the architecture of SAMs, and propose SSPrompt that tackles the identified challenges to learn effective semantic and spatial prompts for SAMs.
Specifically, SSPrompt introduces spatial prompt learning and semantic prompt learning, which optimize spatial prompts and semantic prompts directly over the embedding space and selectively leverage the knowledge encoded in pre-trained prompt encoders.
The two prompt learning methods complement
each other by capturing spatial and semantic information respectively, which together learn effective spatial and semantic prompts for SAMs.
Extensive experiments show that SSPrompt achieves superb image segmentation performance consistently across multiple widely adopted datasets.
Moving forward, we will further explore prompt learning for better prompting SAMs.

{
    \small
    \bibliographystyle{ieeenat_fullname}
    \bibliography{main}
}

\appendix

\section{Dataset Details}
We benchmark our SSPrompt extensively over 6 widely used image segmentation datasets with pixel-wise annotations. As Table 1 (in the main manuscript) shows, the 6 datasets have rich diversity, spanning from street scene data that include high-resolution images captured over different cities and under various daytimes, weathers and seasons, to category-rich data that cover 59 and 150 semantic categories.

\textbf{Cityscapes}~\citep{cordts2016cityscapes} is a dataset designed for visual recognition tasks focused on urban street scenes. This dataset includes a training subset with 2,975 samples and a evaluation subset with 500 samples. Each image in both subsets is annotated at the pixel level, with labels assigned to 19 categories.

\textbf{BDD100K}~\citep{yu2020bdd100k} is a comprehensive dataset tailored for autonomous driving and urban scene analysis. This dataset consists of 7,000 training images and 1,000 validation images collected from various weather conditions, times of day, and urban landscapes, all of which are with pixel-wise annotations of 19 categories.

\textbf{Mapillary}~\citep{neuhold2017mapillary} is a dataset primarily designed for urban scene understanding. This dataset contains 25,000 high-resolution images (e.g., 4000 x 5000) collected from all over the world with pixel-wise annotations. Following prior transfer learning work~\citep{huang2021cross}, we report results over 19 categories shared with Cityscapes.

\textbf{ADE20K}~\citep{zhou2017scene} is a large-scale dataset with 27,574 scene-centric images which consists of 150 categories. This dataset consists of 25,574 training images and 2,000 validation images with pixel-wise annotations.

\textbf{Pascal Context}~\citep{mottaghi2014role} is an extension of PASCAL VOC 2010 detection dataset~\cite{everingham2010pascal}, which contains 59 categories with pixel-wise annotations. It has 4,998 training images and 1,449 validation images.

\textbf{ACDC}~\citep{sakaridis2021acdc} is a dataset designed for robust visual perception. ACDC consists of a large set of 4006 images collected from four common adverse conditions, i.e., fog, nighttime, rain, and snow. For each adverse condition, images are provided with high-quality pixel-level annotation of 19 categories.

\section{More Discussion}

\begin{table}[h]
\caption{Prompt learning of Segment Anything Model, i.e., Segment Anything Model~\citep{kirillov2023segment}. The experiments are conducted on semantic segmentation (in mIoU), where 16-shot data are used (i.e., 16 labelled images for each class) for each dataset.
}
\begin{center}
\resizebox{\linewidth}{!}{
\begin{tabular}{lcccc}
\toprule
Method & SAM~\citep{kirillov2023segment} &CoOp~\citep{zhou2022learning} &LOCN~\citep{parisot2023learning} &\cellcolor{gray!16}{SSPrompt (Ours)}
\\
\midrule
Cityscapes & - &14.3 &14.9 &\cellcolor{gray!16}{20.1} 
\\
\bottomrule
\end{tabular}
}
\end{center}
\label{tab:all_seg_sam}
\end{table}

\textbf{Experiments on Segment Anything Model~\citep{kirillov2023segment}.}
As mentioned in Section 4.5 Discussion in the main manuscript, we did not conduct experiments using SAM~\citep{kirillov2023segment} as its version with text prompt encoder is not open-sourced and the released SAM version can only support class-agnostic segmentation.
Here, we still manage to benchmark with SAM~\citep{kirillov2023segment} to study how the above mentioned issue affects.
Experimental results in Table~\ref{tab:all_seg_sam} show that all prompt learning methods (CoOp~\citep{zhou2022learning}, LOCN~\citep{parisot2023learning} and Our SSPrompt) do not work well, largely because the released SAM version is trained with class-agnostic segmentation and involves little class semantic information.
On the other hand, the results in Tables 2 and 3 in the main manuscript show that all prompt learning methods (CoOp~\citep{zhou2022learning}, LOCN~\citep{parisot2023learning} and Our SSPrompt) achieve much better segmentation performance with SEEM model~\cite{zou2023segment}, because the foundation segmentation model SEEM is trained with both spatial and semantic prompts comprehensively and involves rich class semantic information.

Moreover, experimental results in Table~\ref{tab:all_seg_sam} also show that our SSPrompt outperforms other prompt learning methods~\citep{zhou2022learning,parisot2023learning} clearly on SAM benchmark, indicating that our SSPrompt generalizes better than other prompt learning methods~\citep{zhou2022learning,parisot2023learning}.

Moving forward, we will benchmark our SSPrompt on the recent new semantic-aware SAM~\cite{li2023semantic} such as Semantic SAM and SAM-CLIP~\cite{wang2023sam}, when their codes are released.
\end{document}